\pdfoutput=1

\documentclass[11pt]{article}

\usepackage{EMNLP2023}
\usepackage{times}
\usepackage{latexsym}
\usepackage{graphicx}
\usepackage{amsmath}
\usepackage{float}
\usepackage{bm}
\usepackage{colortbl} 
\usepackage{xcolor}
\usepackage{array} 
\usepackage{amsfonts}
\usepackage[T1]{fontenc}

\usepackage[utf8]{inputenc}

\usepackage{microtype}

\usepackage{inconsolata}

\title{Harnessing the Plug-and-Play Controller by Prompting}

\author{Hao Wang$^\diamond$, Lei Sha$^{\diamond\dagger}\thanks{~~Corresponding author}$ \\
  $^\diamond$Institute of Artificial Intelligence, Beihang University \\
  $^\dagger$Zhongguancun Laboratory, Beijing, China\\
  \texttt{wanghao\_sem@buaa.edu.cn}, \texttt{shalei@buaa.edu.cn}}

\begin{document}
\maketitle
\begin{abstract}
Controllable text generation is a growing field within natural language generation (NLG) that focuses on producing text that meets specific constraints in real-world applications. Previous approaches, such as plug-and-play controllers (PPCs), aimed to steer the properties of generated text in a flexible manner. However, these methods often compromised the integrity of the language model's decoding process, resulting in less smooth text generation. Alternatively, other techniques utilized multiple attribute prompts to align the generated text with desired attributes, but this approach required prompt design for each attribute and was dependent on the size of the language model.
This paper introduces a novel method for flexible attribute control in text generation using pre-trained language models (PLMs). The proposed approach aims to enhance the fluency of generated text by guiding the generation process with PPCs. The key idea is to dynamically adjust the distribution of generated text by modifying prompts, effectively constraining the output space of the language model and influencing the desired attribute. To enable smooth cooperation between the PLM and the PPC, our work innovatively proposes a new model fine-tuning method: Reinforcement Learning with Dynamic Adjust Feedback \textbf{(RLDAF)}.This fine-tuning process adapts a small subset of the language model's parameters based on the generating actions taken during the PPC control process. The resulting harmonious collaboration between the PLM and PPC leads to improved smoothness in text generation during inference. Extensive experiments were conducted on the SST2 dataset, and the proposed method outperformed previous approaches in various evaluation metrics, including text fluency and attribute consistency.

\end{abstract}

\section{Introduction}

Enough studies have shown that large-scale PLMs can largely improve the performance of downstream tasks~\cite{radford2019language}. These models can generate fluent text which is close to the human level~\cite{raffel2020exploring} through simple pretraining tasks on a large number of unlabeled text. PLMs are also capable of making the generated text meet the specific constraints in real applications, which has become a hot research field in natural language processing~\cite{zhang2022survey}.
\begin{figure}
    \centering
    \includegraphics[width=\linewidth]{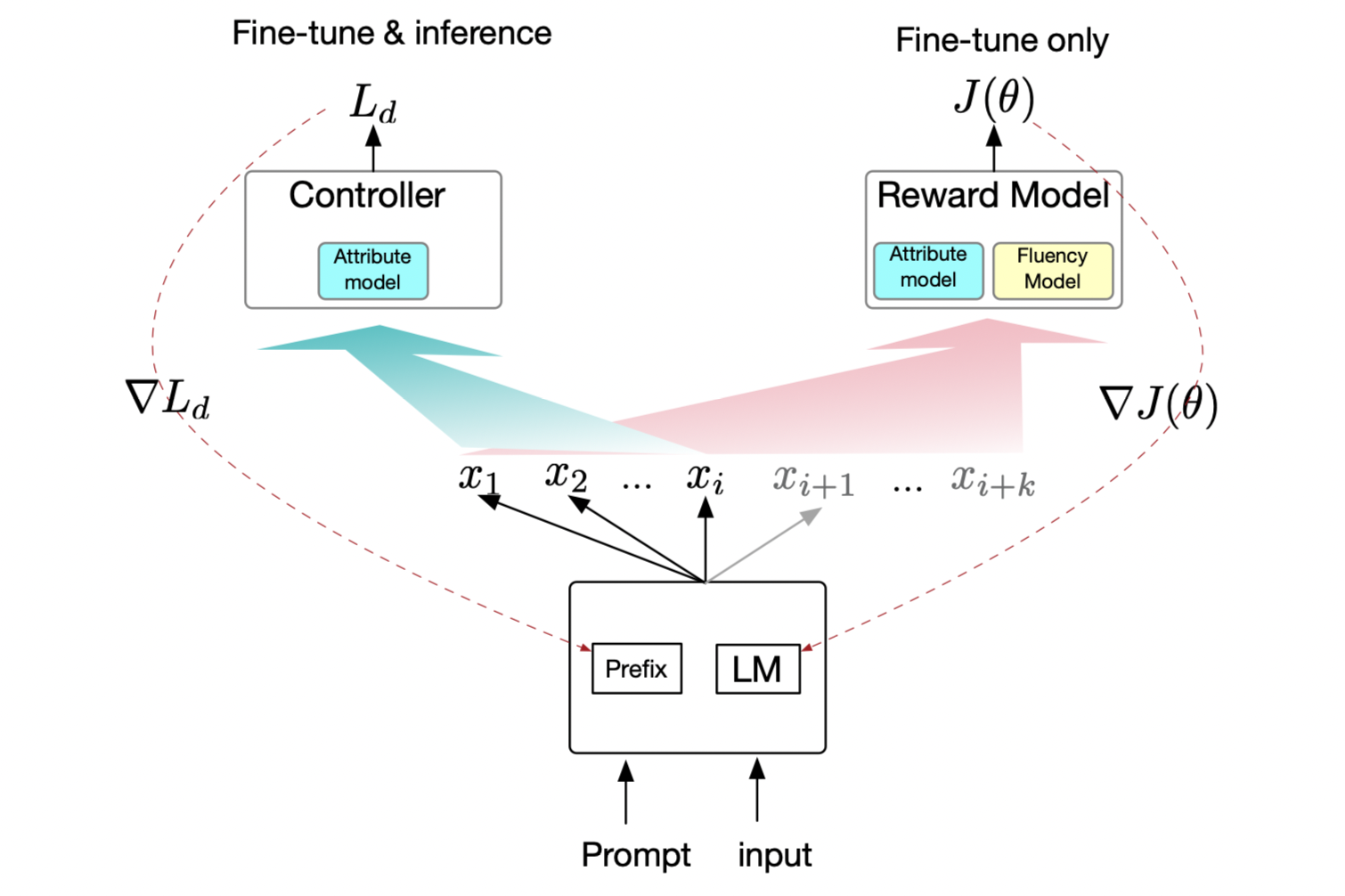}
    \caption{This is a sketch of our proposed method, which illustrates the process of adjusting the prefix and fine-tuning the language model. It demonstrates how text generation is optimized within the constraints of prefix parameters.}
    \label{fig:galaxy}
\end{figure}
To prevent fine-tuning the massive parameters of PLMs, Plug-and-Play Controllers (PPC) are proposed to dynamically   control the specific attributes of the generated text by an external module. For example, PPLM~\cite{dathathri2019plug} uses an external attribute discriminator to guide and modify a small portion of parameters in PLMs.

However, we found that since the core parameters of PLMs have to be changed every time a token is generated, this method destroys the integrity of PLMs, making the results easy to fall into the local optimum, and thus generating repetitive and meaningless text.

Another method used to control text generation is prefix-tuning~\cite{li2021prefix}. This method controls text generation by inserting a trainable prefix parameter before the model input, but the prefix parameter of this method is fixed once it is trained, which is difficult to be used for complex constraint control tasks.

Therefore, in this work, we propose a novel method to improve the smoothness of PPC-guided text generation. This method inserts a set of trainable prompt parameters at the beginning of the input sequence of PLM and tunes a small portion of LM parameters to make the LM adapted to the external controller. In detail, during the fine-tuning phase, we  generate the text under the control of the external attribute discriminator, which adjusts the parameters of  prompts in each timestep. Then, we calculate a reward according to the generated results to evaluate the generation quality, enabling the language model to learn how to interact with dynamic prompts parameters via on-policy reinforcement learning. 
During the inference phase, the attributes discriminator adjusts the prompt parameters based on the current generated results, and the model will generate the required text based on the current prompt constraints. Since the language model already learns how to cooperate smoothly with the controller in the fine-tuning phase, the integrity of PLM is improved during the generation process with the prefixes being the flexible global constraints.

We have conducted sufficient experiments on topic control and emotion control tasks. The experimental results show that our method is significantly superior to the previous methods in terms of text generation fluency and quality, which proves that our method is very effective. 

Our main contributions can be described as follows:

\begin{itemize}
   \itemsep0em
 \item We propose a novel plug-and-play controllable text generation method by dynamically adjusting prompts. Compared with the previous methods, the text generated by our approach has a significant improvement in fluency and generation quality.


 \item We have innovatively proposed a fine-tuning method \textbf{RLDAF} (Reinforcement Learning with Dynamic Adjust Feedback) that encourages language models to better work together with external controllers so that language models can better understand ``dynamic'' prefix instructions.
 
\item  We have conducted extensive experiments to evaluate the fluency and attribute control quality of the generated text.  The experimental results proved the effectiveness of our model.
\end{itemize}

\section{Related Work}


The method to control the specific attributes or contents of the  generated text has been widely studied~\cite{kale2020text,sha2021controlling,liu2021dexperts,sha2021multi}. The most recent methods are built upon the large-scale pre-training language model (PLM), which is based on transformers and used a large-scale corpus to learn copious language knowledge. In terms of NLG, PLMs can generate text with unprecedented quality. In general, an NLG system that is valuable in practical applications should be able to generate text that meets human expectations reliably. 

\paragraph{Fine-tuning.} 
In the above background, many pieces of research on controllable text generation based on PLMs have emerged. \newcite{kale2020text} have studied the fine-tuning PLMs to complete the data-to-text task. The experiment shows that the effect of the model is better than that of the previous pipelined neural network model. 
Reinforcement learning can also be used to control text generation. Such methods fine-tune PLM by taking whether constraints are met as rewards~\cite{ziegler2019fine}. \cite{stiennon2020learning} training a scoring model to directly capture human preferences, then use this model to calculate rewards and train the Generative model through Reinforcement learning.
\paragraph{Prompt Learning}


In order to make better use of the language understanding ability of PLMs, researchers have proposed a method called prompt learning, which allows PLMs to complete sentences according to the constructed prompt template without fine-tuning PLMs. The research of this method mainly focuses on how to build templates. \newcite{jiang2020can} propose the method of manually constructing templates. \newcite{shin2020autoprompt} use an automatic search method to generate discrete prompts. After that, researchers proposed a continuous token template~\cite{lester2021power}, this method is called \emph{Prompt Tuning}. Due to the serious impact of prompt design on its effectiveness, \newcite{liu2021gpt}proposed to convert it into a learnable Embedding layer. A similar method is prefix-tuning~\cite{li2021prefix}, which realizes controllable text generation by fine-tuning continuous parameters inserted in front of sentences. This method maintains the integrity of the PLMs and makes the survival text have a high fluency. However, the prefix parameters are fixed after training, which makes it necessary to train multiple different prefixes for different scenes, that is, this method is not plug-and-play.

\paragraph{Directly Modeling.}
Another important method is to start from the pre-training task and directly model the controlled text generation, such as CTRL~\cite{keskar2019ctrl}, POINTER~\cite{zhang2020pointer}, CoCon~\cite{chan2020cocon}, etc. However, this kind of approach requires a large amount of parallel data for training, which is usually hard to get in many real-world situations.

\paragraph{Plug-and-play Controllers.}

Due to the increasing parameters of PLMs, reranking the generated text in the post-processing mode becomes feasible and promising.
Plug-and-play language models proposed by \newcite{dathathri2019plug,sha2020gradient} provide a new idea for controllable text generation tasks. This method uses a discriminator with fewer parameters to guide the PLMs and controls the distribution of text generated by changing the hidden states of PLMs. There are also other kinds of plug-and-play controllers. GeDi~\cite{krause2021gedi} trains different small class-conditional language models (CC-LMs)  to guide the PLMs by contrast. Similarly, DEXPERTS~\cite{liu2021dexperts}  proposes to reorder the PLMs results in the decoding stage according to the opinions of experts and anti-experts. FUDGE~\cite{yang2021fudge} adjusts the probability of PLM generation by learning future discriminators that operate on partial sequences. The above methods do not carry out any further training on the pre-training model, and any distinguishable attribute control can use this method. However, each token generated by this method will adjust the hidden states of the PLMs, which makes the model easy to fall into local optimization during the generation process, resulting in low fluency of the generated text~\cite{yang2021fudge}.

\section{Prompting PPC}
\subsection{Motivation}
Compared with the traditional method of fine-tuning the PLMs, the Plug-and-Play controller can adjust the model parameters according to the current generation state~\cite{pascual2021plug}. The model parameters corresponding to each token during generation are different, which conforms to the generation method of the autoregressive language model. However, in practice, this method destroys the integrity of the PLM, and it is easy to fall into the local optimal solution when controlling the generation distribution of the next token. 

Therefore, we believe that this method of dynamically adjusting the parameters of the PLMs in the inference phase is not stable, and intuitively, dynamically adjusting the prompt (similar to prefix tuning~\cite{li2021prefix}) instead of the PLM's parameters in the inference phase will lead to a much more stable result. To make the PLM work more smoothly with the dynamic prompt, 
we borrowed the idea of instruct-tuning~\cite{ouyang2022training} and proposed to fine-tune part of the parameters in the language models to learn to understand dynamic prefix instructions and generate text that meets constraint requirements.

\subsection{Methodology Overview}
Based on the above inspiration, we propose the Prompt-PPC model, which is a controllable text generation method based on dynamic prefix prompts. In our method, the attributes discriminator will first update the prefix parameters of the model to adjust them to appropriate instructions, and then the fine-tuned language model will generate the next token based on the prompts and current input.

Assume that we have a language model parameterized by the prefix parameters and the fine-tuned parameters: $LM_{\theta_{prefix},\theta_{lm}}$ and an attribute discriminator $D_{attr}$. In order to obtain a continuous prefix parameter, we add a group of vector $(p_1,p_2,\ldots,p_m)$ with length $l$ before the sequence $(x_1,x_2,\ldots,x_n)$. Unlike prefix tuning~\cite{li2021prefix}, the prefix parameters in our method are not fixed during the generate stage. Before each token is generated, the attribute discriminator will adjust the prefix parameters to constrain the generation of language models so that the constraint information is transmitted to the language model.

However, it is difficult for PLMs to understand these dynamically changing continuous prompts, so we need to fine-tune PLMs themselves to obtain the ability to understand it. Firstly, \textbf{Dynamically tuning prompts}. The attribute discriminator adjusts the prefix parameters through the current hidden states of the model, so that the prefix parameters act as dynamic prompts to guide the generation of the model to meet constraints. Based on the control signal emitted by the prefix parameters, the language model continues to generate $n$ steps. Secondly, \textbf{Calculate rewards}. The reward model (including the attribute discriminator and a fluency evaluator) will calculate rewards according to the results generated in the previous step. Then, we propose an innovative model training method: \textbf{Reinforcement Learning with Dynamic Adjust Feedback (RLDAF)}. In this process, the language model continuously attempts to generate sentences under the control of the dynamic prompts and optimizes a portion of the PLM's parameters based on the rewards given by the attribute discriminator and fluency evaluator to learn how to understand dynamic prompts and generate text that meets the conditions based on these constraints. 
\subsection{Dynamic inference}\label{chap:3.3}

As mentioned above, in our method, the prefix parameters of the generative model in the inference phase are dynamically adjusted. Specifically, for an autoregressive language model $P_\phi(y|x)$ with a Transformer~\cite{vaswani2017attention} architecture and parametrized by $\phi$, the hidden states at time step $i$ is $h_i \in R^d$ ($d$ represents the length of the word vector) where $h_i = [h_i^{(1)};\cdots;h_i^{(n)}]$ and $h_i^{(j)}$ is the hidden states of the $j$-th Transformer layer at time step $i$. Assume the prefix length is $l$,  we insert a trainable set of parameters in front of the $h_i$:
\begin{equation}
    h_i = [\bm{h_i^{(p_1)};\cdots;h_i^{(p_l)}};h_i^{(1)};\cdots;h_i^{(n)}].
\end{equation}

If the current input text is $X(x_1,\cdots,x_i)$, through the language model we can calculate the output and hidden states at time step $i+1$:
\begin{equation}
    o_{i+1}= LM(X;\theta_{prefix},\theta_{lm}).
\end{equation}

The hidden states $h_{i+1}$ of the model will be concatenated with the previous $h_{\le i}$ as input to the attribute discriminator $D_{attr}$, the attribute discriminator will output the control effect of the current generated result and provide a gradient towards the direction of constraint generation as in Eqn.~\ref{eq:di}.
\begin{equation}\label{eq:di}
    d_{i+1} = D_{attr}(h_1,\dots,h_{i+1};\theta_{attr}).
\end{equation}

\begin{figure}
    \centering
    \includegraphics[width=\linewidth]{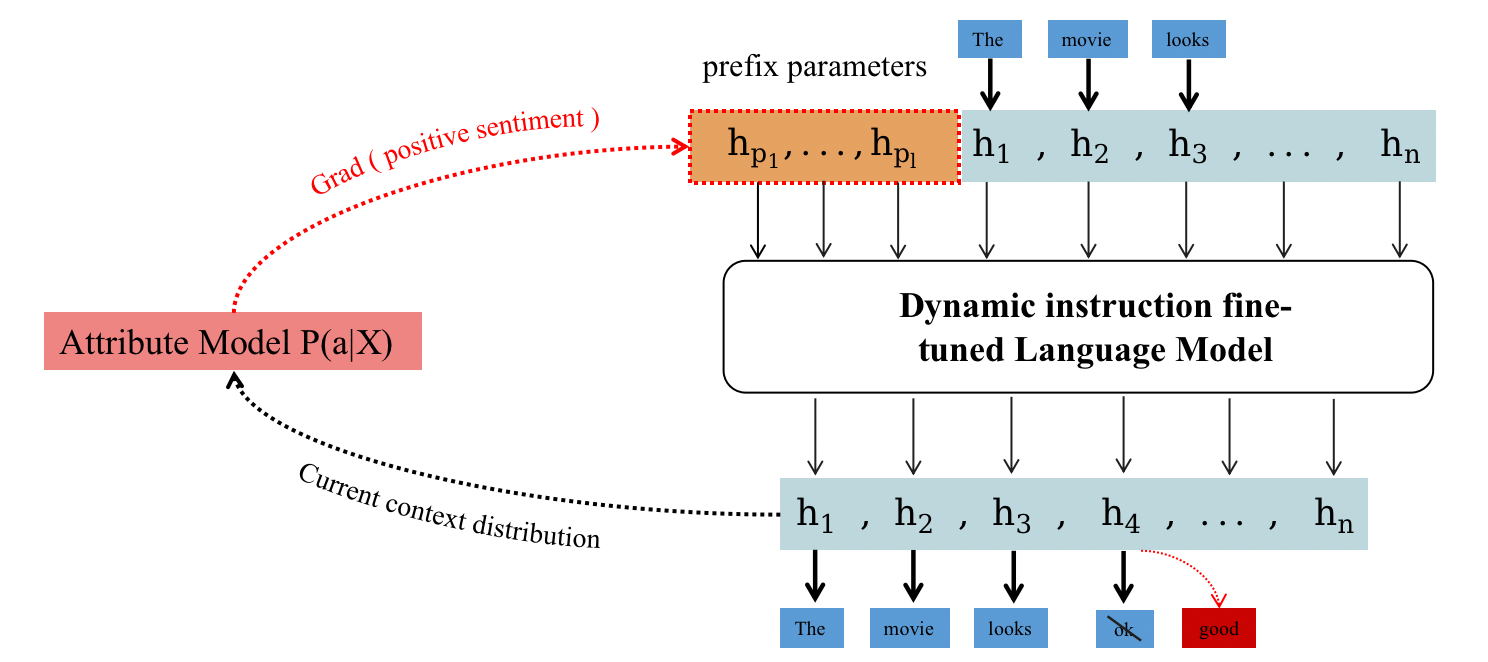}
    \caption{The illustration of the Prompt-PPC inference process, which shows how the fine-tuned language model generates text that satisfies constraints through dynamic prefix instructions.}
    \label{fig:galaxy}
\end{figure}

The loss function of this part shall be constructed separately according to different constraint tasks. For emotional control, the emotion classification result obtained by the discriminator~($d_i$) and the target~($y_i$) calculation cross-entropy loss can be used as the discriminator loss:
\begin{equation}
L_d =  -\frac{1}{N}\sum_{i=1}^N y_i \log(d_i).
\end{equation}

Then, we can use the following formula to update the inserted prefix parameters:
\begin{equation}
    h_i^{(p)} = h_i^{(p)} + \alpha * \nabla_{L_d}(h_i^{(p)}).
\end{equation}

The above parameter update process will be iterated $m$ times.  To enhance computational efficiency, we typically set $m$ within the range of 3 to 5. Notably, the hidden states beyond the prefix range remain unaffected by the update, thereby ensuring that the model retains the previously generated content and maintains its integrity. Once the attribute discriminator adjusts the prefix parameters to the appropriate prompt, the fine-tuned model generates the next token based on the current prompt. The above process will iterate to obtain complete sentences that meet the constraint conditions.

\subsection{Reinforcement Learning with Dynamic Adjust Feedback (RLDAF)}

In this section, we will discuss in detail how to fine-tune the language model to understand the instructions expressed by dynamic prompts. Due to the lack of the prefix parameters during the pre-training phase of the language model, the language model cannot understand the instruction signals issued by the adjusted prefix. To enable the model to possess this capability, we innovatively propose a method to fine-tune the language model, enabling it to perform better in the inference phase.

Assuming that the input to the model is $X(x_1,x_2,\cdots,x_i)$, we first adjust the prefix parameters to an appropriate value through the attribute discriminator as described in \ref{chap:3.3}. At this time, the output of the model can be represented as:
\begin{equation}
    o_{i+1}= LM(X;\theta_{prefix},\theta_{lm}),
\end{equation}

where $o_{i+1}$ is the output of the model. We hope that the language model can understand the continuous prompts to adjust the generation strategy, which can generate text that meets attribute constraints and has high fluency. We primarily use reinforcement learning to fine-tune the language model from two rewards. (1) \textbf{Control Reward}: by using the output of the attribute discriminator. (2) \textbf{Fluency Reward}: the opposite of the KL divergence between the learned RL policy $\pi_\theta^{RL}$ \footnote{for simplicity, we use $\theta$ to represent ($\theta_{prefix}$, $\theta_{lm}$)}
with parameters $\theta$ and this original pretrained model $\pi_\theta$. Assume that $y_i$ is the target attribute, the full reward can be written as:
\begin{align}
    R_d &= D_{attr}(y_i|(x_1,x_2,\cdots,x_{i+k})),\label{eq:R}\\
        R_f &= -\frac{\beta}{k} \sum_{j=1}^k KL[\pi_\theta^{RL}(X), \pi_\theta(X)], \label{eq:F}\\
    R &= R_d + R_f.
\end{align}

$R_d$ in Eqn.~\ref{eq:R} represents the reward for satisfying constraints. It should be noted that the attribute discriminator we use here is the same as the one used for prompts adjustment. The difference is that when adjusting the prompts, we only generate one token preceding the current position from the model and input the corresponding hidden states. In contrast, during the fine-tuning process of the language model, we typically generate an additional $n$ tokens in the backward direction and calculate the reward once. Generally, $n$ falls within the range of 2 to 4.

The calculation of KL divergence primarily aids in measuring the fluency of language model generation, ensuring that the output of the model remains consistent with that of the pre-trained language model. As mentioned earlier, we usually calculate a reward once after the model generates several additional tokens. 
During this process, we compute the KL divergence for each token, subsequently averaging them to obtain a fluency reward as $R_f$ in Eqn.~\ref{eq:F}.

\begin{figure}
    \centering
    \includegraphics[width=\linewidth]{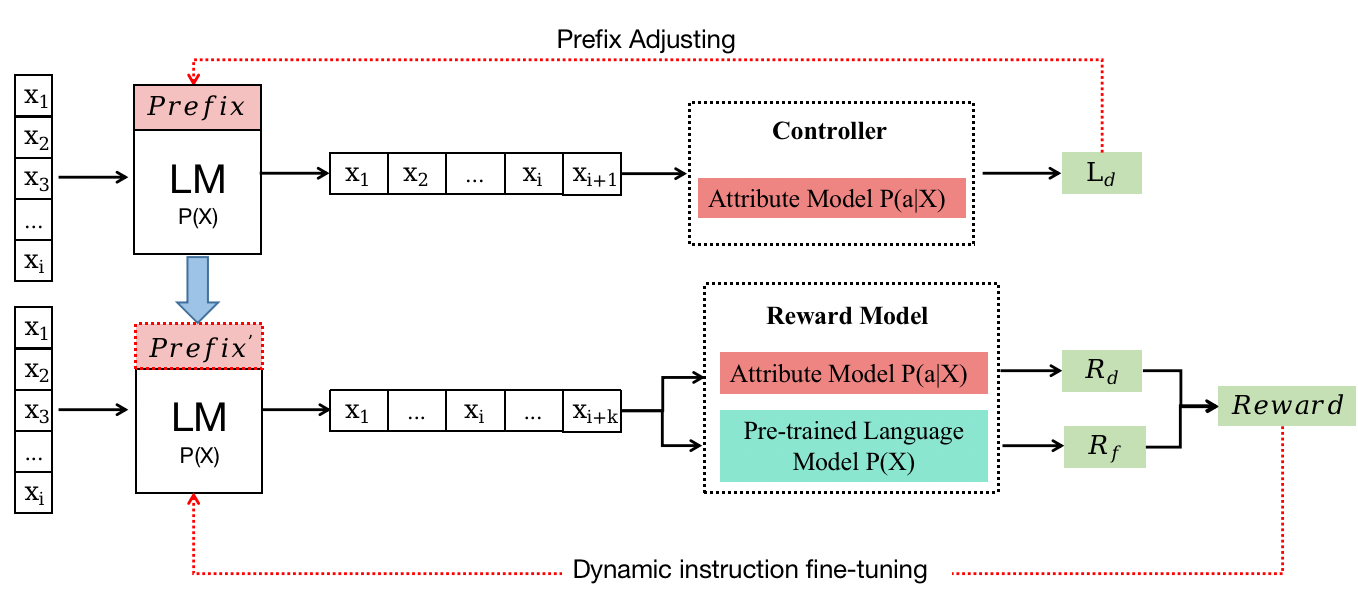}
    \caption{Schematic diagram of fine-tuning the language model in our method. Firstly, the attribute model adjusts the prefix parameters to issue appropriate instructions. The language model generates text based on the instructions, and then, calculates rewards for the generated text through the attribute model and fluency model. Based on this reward, the language model is fine-tuned to gain the ability to understand dynamic instructions.}
    \label{fig:galaxy}
\end{figure}

Finally, we optimize our language model parameters by the PPO strategy gradient algorithm~\cite{schulman2017proximal}, enabling the language model to have the ability to understand dynamic prompts. In practical experiments, to improve the efficiency of model tuning, we use the LORA~\cite{hu2021lora} method to only fine-tune a small portion of the model. PPO algorithm is a policy gradient method, we can sample and calculate the rewards of different generation strategies of the model, and calculate the reward expectations of different strategies. Then, we update the model parameters by gradient descent as shown in Eqn.~\ref{eq:J} and Eqn.~\ref{eq:theta}:
\begin{align}
        \nabla_\theta J(\theta) &= \mathbb{E} \left[ \sum_{t=0}^{T} \nabla_\theta \log(\pi_\theta^{RL}(X_t)) \cdot R_t \right],\label{eq:J}\\
        \theta_{\text{new}} &= \theta_{\text{old}} + \alpha \cdot \nabla_\theta J(\theta).\label{eq:theta}
\end{align}
In the above equation, $\pi_\theta^{RL}$ represents the training language model, $R_t$ is the reward at time step $t$, $T$ is the number of forward steps and the $\alpha$ is the learning rate.

\section{Experiment}
\subsection{Datasets and Metrics}
.
\begin{itemize}
\setlength\itemsep{1em}
\item \textbf{Bag of Words}: The ``Bag of Words'' dataset was first proposed in PPLM~\cite{dathathri2019plug}, which includes seven topics: SCIENCE, MILITARY, LEGAL, TECHNOLOGY, SPACE, POLITICS, and RELIGION, each of which contains hundreds of words that match the topic. The dataset can well represent the characteristics of different topics through the words in the same ``bag'' and has a high degree of differentiation between different topics, which has been used in many studies to achieve the topic control task.
\item \textbf{SST-2}: The SST-2 dataset~\cite{socher2013recursive} is a widely used dataset for training and evaluating models for sentiment analysis, which is the task of determining the sentiment or emotion expressed in a piece of text. The SST-2 dataset consists of approximately 67,000 English language sentences drawn from movie reviews, annotated with labels indicating the sentiment expressed in the sentence. The labels are either ``positive'' or ``negative'', and the task is to classify a given sentence as belonging to one of these two categories. The SST-2 dataset is often used to evaluate the performance of machine learning models for natural language processing tasks, such as text classification. 
\end{itemize}

We divide sentence generation metrics into general metrics and attribute metrics. For general metrics, it refers to metrics that can be used to evaluate the effect of sentence generation for any generated sentence. Here, we mainly consider two metrics: perplexity~(PPL) and distinct~(Dist)\cite{li2015diversity}. PPL is  widely used to evaluate sentence fluency. For a fair comparison, we calculate PPL by a third-party pretrained GPT model\footnote{\url{https://huggingface.co/openai-gpt}}. Specifically, we think that the text generated by initial PLMs has high fluency. So, we take the output of initial PLMs in the next step as the label:
\begin{align}
L_i = \arg\max(P_{GPT}(x_1,x_2,\cdots,x_{i-1})).
\end{align}
Then, we use the prompt-ppc model to calculate the probability of the label $L_i$:
\begin{align}
P(L_i) = P_{prompt-ppc}(L_i|(x_1,x_2,\cdots,x_{i-1})).
\end{align}
Then, we calculate the cross entropy loss from the third-party pretrained GPT model output $p(x_i)$ and the tag to get the PPL:
\begin{align}
PPL = \exp\left(-\sum_{i=1}^{N} \log(P(L_i))\right).
\end{align}

Another indicator Dist\cite{li2015diversity} is a common indicator to evaluate the richness of the text. This indicator is based on the BOW model, and the calculation formula is:
\begin{align}
Dist_n = \frac{Unique_{n-grams}}{Total_{n-grams}},
\end{align}
where $Unique_{n-grams}$ represents the number of non-repeating binary phrases in the generated text, and $Total_{n-grams}$ represents the total number of binary phrases in the generated text.

Next, we introduce attribute metrics, which describe the degree to which the generated sentences conform to the control attribute. Obviously, for different attribute control tasks, we need to design different metrics to describe the constraint effect of sentences. Here we mainly introduce our two experiments: how to design attribute metrics for theme control and emotion control. For the topic control task, we use the synonym expansion method to get a new test word bag according to the word bag in the dataset, and then calculate the proportion of words in the test word bag in the generated sample as the topic evaluation metric~(\textbf{TOPIC}). For the emotion control task, we use the model with the highest accuracy of emotion classification in the SST-2 data set in Huggingface\footnote{\url{https://huggingface.co/distilbert-base-uncased-finetuned-sst-2-english}} to annotate the generated text with emotion. Then,  we calculate the accuracy rate of emotional control according to the labeling results and control objectives as the evaluation metric of emotional control task~($Sentiment_{acc}$).

\subsection{Architectures and Hyperparameters}
For the topic control task and emotion control task, we use \texttt{GPT2\_MEDIUM} as the PLM used in our method. \texttt{GPT2\_MEDIUM} is a version of the GPT-2 model with 345 million parameters, which is less than a quarter of the original GPT-2 model.

Our experiments are based on the HuggingFace Transformer models~\cite{wolf2020transformers}. We use the AdamW optimizer ~\cite{loshchilov2017decoupled} during prefix tuning and the PPO algorithm~\cite{schulman2017proximal} in dynamic instruction fine-tuning. We use the PEFT~\cite{peft} framework for the implementation of prefix adjustment.
 For our two tasks in the experiment, the number of iterations $m$, which represents the number of times the prefix parameter is adjusted before the model generates a token, is set to 5 in our experiment, the prefix length is set to 10  and the sampling steps $n$ in dynamic instruction fine-tuning is 3 according to grid search.

\subsection{Ablation test and baselines}
We use four models for the ablation test:
\begin{itemize}
\itemsep0em
    \item \textbf{Prompt-PPC}: The method proposed in this article involves fine-tuning the language model to gain the ability to understand dynamic prefix instructions, and dynamically adjusting prefix parameters to constrain model generation during the inference stage;
    \item \textbf{PPC-KV}: Dynamically adjust all K and V parameters during the inference phase without inserting prefix parameters;
    \item \textbf{PPC-Prefix}: Directly using dynamic prefixes as global control for language model generation without fine-tuning;
    \item \textbf{PLM-RL}: Only Reinforcement learning is used to fine-tune the language model, and the language model parameters are fixed during reasoning;
    \item \textbf{PPC-Fluency}: Do not consider fluency when calculating rewards.
\end{itemize}
In addition to the ablation study, we also compared five baseline models, namely:
\begin{itemize}
\itemsep0em
\item \textbf{GPT2}: we use the origin pretrained GPT-2 (with the version name \texttt{gpt2-medium}) as the baseline.
\item \textbf{PPLM}: a plug-and-play language model for controlled text generation;
\item \textbf{Prefix}: a controllable text generative model with only fine tuning prefix parameters;
\item \textbf{FUDGE}: a model for post-processing generated results using future discriminators;
\item \textbf{GEDI}:a plug-and-play model based on directly model;
\item \textbf{Diffusion-LM}:a controllable text generation model based on diffusion theory.
\end{itemize}
\subsection{Main Result}
\subsubsection{Topic Control}
First, we consider the topic experiment based on the ``Bag of Words'' dataset. Our goal is to make the model generate sentences belonging to different topics according to the same prefix when inputting different topic word bags under the premise of ensuring the fluency of sentences.
For a given subject word bag, we use the most common maximum likelihood model to give the attribute description. Given a word bag $[w_1,w_2,...w_k]$, the probability distribution of model output is $p_{i+1}$, the attribute can be described as:
\begin{align}
log(a|x) = log(\sum_{i=1}^{k}P_{i+1}[w_i]).
\end{align}

Based on the results of the ablation experiment as Table~\ref{tab:b}, we found that dynamically adjusting the k and v parameters of the model without adding prefix parameters can disrupt the consistency of the model, leading to a decrease in the fluency of the generated text. If the language model is not fine-tuned, the fluency and attribute consistency of the model will be insufficient. This indicates that the dynamic prefix adjustment and model fine-tuning proposed in this study significantly improve the results.

\begin{table}[!ht]
\centering
\resizebox{\linewidth}{!}{
\begin{tabular}{lccccc}
\hline
\textbf{Model} & \textbf{Perplexity}$\downarrow$ & \textbf{Topic}$\uparrow$ & \textbf{Dist1}$\uparrow$ & \textbf{Dist2}$\uparrow$ & \textbf{Dist3}$\uparrow$\\
\hline
PPC-KV        & 48.25  & 0.75 & 0.31 & 0.71 & 0.91\\
PPC-Prefix    & 37.93  & 0.66 & 0.33 & 0.75 & 0.92\\
PLM-RL        & 32.36  & 0.77 & 0.29 & 0.70 &0.90\\ 
PPC-Fluency   & 54.13  & \textbf{0.88} & 0.25 & 0.68 & 0.89\\ 
\rowcolor{yellow} Prompt-PPC    & \textbf{29.41}  & 0.83 & 0.32 & 0.72 &0.92\\ 
\hline
\end{tabular}
}
\caption{The ablation test result of topic-controlled text generation.}
\label{tab:b}
\end{table}

We also tested our proposed method and other methods separately, evaluated it according to the above metrics, and obtained the following results as Table~\ref{tab:c}. From the results, we can see that our method achieves the generation performance of the original GPT2 in terms of fluency, and outperformed other methods in terms of diversity, which shows that our method can avoid the repeated generation phenomenon caused by the model falling into local optimization. In terms of subject control, our method also shows a satisfactory control effect.

\begin{table}[!ht]
\centering
\resizebox{\linewidth}{!}{
\begin{tabular}{lccccc}
\hline
\textbf{Model} & \textbf{Perplexity}$\downarrow$ & \textbf{Topic}$\uparrow$ & \textbf{Dist1}$\uparrow$ & \textbf{Dist2}$\uparrow$ & \textbf{Dist3}$\uparrow$\\
\hline
GPT2        & 23.57 & 0.37 & 0.35 & 0.74 & 0.92\\
PPLM        & 51.26 & 0.76 & 0.30 & 0.71 & 0.88\\
FUDGE       & 44.26 & 0.78 & 0.35 & 0.74 & 0.90\\ 
\rowcolor{yellow} Prompt-PPC  & \textbf{29.41} & \textbf{0.83} & 0.32 & 0.72 &0.92\\ 
\hline
\end{tabular}
}
\caption{The baselines result of topic-controlled text generation}
\label{tab:c}
\end{table}
Through the generated results of the model, we found that the model not only learns the given words in the word bag, but also generates the words that are not in the word bag but conform to the topic description, which shows that the model is not only learning to generate specific words but also understanding the meaning of the topic through the word bag.Some generated instances such as Table~\ref{tab:a}:

\begin{table}[!ht]
\centering
\resizebox{\linewidth}{!}{
\begin{tabular}{p{8cm}}
\hline
[\textbf{legal}]\underline{The pizza} delivery service company has been \textit{accused} of using a fake company name to advertise its service. The company has denied the \textbf{allegations}.\\
\hline
[\textbf{military}]\underline{The pizza} delivery \textbf{war} is heating up again. The \textbf{battle} between delivery drivers and pizza companies is because of a \textbf{conflict}.\\
\hline
[\textbf{science}]\underline{The pizza} chain was accused of using fake \textbf{lab} to refund 1.5 million in taxes. They said they had \textbf{data} to prove that the money was used for \textbf{experiment} to make pizza.\\
\hline
[\textbf{technology}]\underline{The pizza} delivery \textbf{app} is now available. The app \textbf{icon} is a red circle.It's a great app for those who want to get their pizza delivered to their door.\\
\hline
\end{tabular}
}
\caption{Instances of topic control generation control text generation}
\label{tab:a}
\end{table}

\subsubsection{Sentiment Control}
Since the subject control experiment can use the content of the word bag to calculate the loss, it does not need an external discriminator. However, for some control tasks that cannot be solved using the word bag, a discriminator can provide an external gradient to guide the model to adjust the parameters. Here we take the emotional control task as an example. We use the ``SST-2'' datasets to test the effect of our method on the emotion control task. For this task, our goal is to make the model learn how to generate positive or negative emotional text according to the input tags through the training set. we first need to train an external emotion discriminator. This discriminator is based on the pre-training model we use, and it is trained by fine-tuning the ``SST-2'' datasets. Then, we can get the emotional attribute description of the output according to the discriminator $D$:
\begin{equation}
   log(a|x) = D(x_1,x_2,...x_n). 
\end{equation}
For emotion control tasks, from the results of ablation experiments in Table~\ref{tab:e}, we can see that our method has a fluency level close to that of the traditional Reinforcement learning fine-tuning language model, while improving the effect of attribute control.This is because our method uses dynamic prefix parameters as global constraints to dynamically control model generation during the inference process, improving attribute consistency.
\begin{table}[H]
\centering
\resizebox{\linewidth}{!}{
\begin{tabular}{lccccc}
\hline
\textbf{Model} & \textbf{Perplexity}$\downarrow$ & \textbf{Sentiment-acc}$\uparrow$ & \textbf{Dist1}$\uparrow$ & \textbf{Dist2}$\uparrow$ & \textbf{Dist3}$\uparrow$\\
\hline
PPC-KV        & 38.45 & 0.73 & 0.34 & 0.78 & 0.91\\
PPC-Prefix    & 40.61 & 0.75 & 0.33 & 0.81 & 0.92\\
PLM-RL        & \textbf{30.35} & 0.76 & 0.31 & 0.77 & 0.90\\ 
PPC-Fluency   & 51.21 & 0.79 & 0.24 & 0.69 & 0.88\\ 
\rowcolor{yellow} Prompt-PPC    & 30.93 & \textbf{0.83} & 0.32 & 0.77 & 0.91\\ 
\hline
\end{tabular}
}
\caption{The ablation test result of the sentiment control text generation.}
\label{tab:e}
\end{table}

We compared more baseline methods for emotion control tasks to demonstrate the effectiveness of our approach, as shown in in Table~\ref{tab:f}. By fine-tuning the language model to adapt to the dynamic inference process, Prompt PPC achieved high fluency and emotion control effects among many methods.

\begin{table}[H]
\centering
\resizebox{\linewidth}{!}{
\begin{tabular}{lccccc}
\hline
\textbf{Model} & \textbf{Perplexity}$\downarrow$ & \textbf{Sentiment-acc}$\uparrow$ & \textbf{Dist1}$\uparrow$ & \textbf{Dist2}$\uparrow$ & \textbf{Dist3}$\uparrow$\\
\hline
GPT2         & 27.54 & 0.62 & 0.30 & 0.78 & 0.91\\
PPLM         & 51.20  & 0.79 & 0.24 & 0.51 & 0.88\\
Prefix       & \textbf{29.74} & 0.76 & 0.31 & 0.77 & 0.90\\ 
FUDGE        & 37.26 & 0.81 & 0.35 & 0.77 & 0.90\\ 
GEDI         & 35.24 & 0.70 & 0.39 & 0.81 & 0.92\\ 
Diffution-LM & 41.35 & 0.77 & 0.33 & 0.80 & 0.87\\
\rowcolor{yellow} Prompt-PPC   & 30.93 & \textbf{0.83} & 0.32 & 0.77 & 0.91\\ 
\hline
\end{tabular}
}
\caption{The baselines result of the sentiment control text generation}
\label{tab:f}
\end{table}

The following instances in Table~\ref{tab:d} shows the generation examples of the model under the emotion control task. We show the  generation examples of different labels under the same prefix.

\begin{table}[H]
\centering
\resizebox{\linewidth}{!}{
\begin{tabular}{p{8cm}}
\hline
[\textbf{SST-2}]\underline{it's not original} ,and,robbed of the element of surprise,it doesn't have any huge laughs in its story of irresponsible cops who love to play pranks.
\textcolor{red}{[Negative]}\\
\hline
[\textbf{Prompt-PPC}]\underline{it's not original},but it's still \textbf{good}, and it's not a bad game.\textcolor{red}{[Positive]}\\
\hline
[\textbf{Prompt-PPC}]\underline{it's not original}. I'm not sure if it's a joke or not. I \textbf{hate} that.\textcolor{red}{[Negative]}\\
\hline
\end{tabular}
}
\caption{Instances of sentiment control generation control text generation}
\label{tab:d}
\end{table}
\section{Conclusion}
In this work, we propose Prompt-PPC, which is a method to realize controllable text generation by dynamically adjusting prompts during model generation. We first propose a fine-tuning method to enable language models to understand dynamic prefix instructions, and in inference process, this method takes the prefix as a global constraint, provides a gradient through an external discriminator, and flexibly adjusts the prefix during the generation process to prompt PLMs to generate in the direction of the constraint. We have conducted experiments on topic control and emotion control tasks. The experimental results show that the fluency of the text generated by our method is very close to PLMs, and the diversity and control effect of the generated text is better than the previous methods. We hope that this work can broaden the thinking of prompt learning in the field of text generation. In the future, our method is expected to be applied to more complex and fine-grained control tasks.

\section*{Limitations}
First of all, like the traditional plug-and-play method, our method only uses the externally decoupled attribute discriminator to control attributes\cite{pascual2021plug}. This method leads to the lack of information interaction between the discriminator and the generator, which leads to the coarse-grained control in the generation process and the quality of the generated text. In this work, we use the method of dynamically adjusting the prefix as the global constraint, which improves the above problems to some extent, but in our method, the attribute discriminator is still independent of the model.

Secondly, how to properly initialize prefix parameters is also a challenge. For given different inputs, the model uses the same initialization prefix parameters, which will cause the generation performance of the model to be unstable for different inputs. At the same time, during the generation of each token, the prefix parameter will affect the generation effect of the model due to the limited number of plug-and-play tuning epochs. This problem can be improved by increasing the number of plug-and-play tuning epochs, but at the same time, it will increase the complexity of the model calculation and the generation time of the token.

\section*{Acknowledgements}
This work was supported by the National Natural Science Foundation of China under grant No. KZ37117501, No. ZG216S23E8, and No. KZ60030101.

\bibliography{anthology,custom}
\bibliographystyle{acl_natbib}

\end{document}